\documentclass{article}

\usepackage{arxiv}
\usepackage[utf8]{inputenc}
\usepackage[T1]{fontenc}   
\usepackage{hyperref}       
\usepackage{url}           
\usepackage{booktabs}      
\usepackage{amsfonts}       
\usepackage{amsmath}
\usepackage{nicefrac}      
\usepackage{microtype}      
\usepackage{lipsum}		
\usepackage{graphicx}
\usepackage{caption}
\usepackage{subcaption}
\usepackage{natbib}
\usepackage{xcolor}
\usepackage{doi}
\usepackage{algorithm}
\usepackage{algpseudocode}
\usepackage{amsthm}

\newcommand{\bm}[1]{\mathbf{#1}}

\title{DKL-KAN: Scalable Deep Kernel Learning using Kolmogorov-Arnold Networks}

\author{{\includegraphics[scale=0.06]{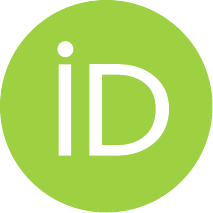}\hspace{1mm}Shrenik Zinage}\footnote{Corresponding author} \\
	School of Mechanical Engineering\\
	Purdue University\\
	West Lafayette, IN \\
	\texttt{szinage@purdue.edu} \\
    \And
    {\includegraphics[scale=0.06]{orcid.pdf}\hspace{1mm}Sudeepta Mondal} \\
	RTX Technology Research Center, \\
	411 Silver Lane,\\
	East Hartford, CT \\
	\texttt{sudeepta.mondal2@rtx.com} \\
    \And
    {\includegraphics[scale=0.06]{orcid.pdf}\hspace{1mm}Soumalya Sarkar} \\
	RTX Technology Research Center, \\
	411 Silver Lane,\\
	East Hartford, CT \\
	\texttt{soumalya.sarkar@rtx.com} \\
}

\renewcommand{\shorttitle}{DKL-KAN}
\hypersetup{
  pdftitle={DKL-KAN},
  pdfauthor={Shrenik Zinage, Sudeepta Mondal, Soumalya Sarkar}
}

\begin{document}
\maketitle

\begin{abstract}

The need for scalable and expressive models in machine learning is paramount, particularly in applications requiring both structural depth and flexibility. Traditional deep learning methods, such as multilayer perceptrons (MLP), offer depth but lack ability to integrate structural characteristics of deep learning architectures with non-parametric flexibility of kernel methods. To address this, deep kernel learning (DKL) was introduced, where inputs to a base kernel are transformed using a deep learning architecture. These kernels can replace standard kernels, allowing both expressive power and scalability. The advent of Kolmogorov-Arnold Networks (KAN) has generated considerable attention and discussion among researchers in scientific domain. In this paper, we introduce a scalable deep kernel using KAN (DKL-KAN) as an effective alternative to DKL using MLP (DKL-MLP). Our approach involves simultaneously optimizing these kernel attributes using marginal likelihood within a Gaussian process framework. We analyze two variants of DKL-KAN for a fair comparison with DKL-MLP: one with same number of neurons and layers as DKL-MLP, and another with approximately same number of trainable parameters. To handle large datasets, we use kernel interpolation for scalable structured Gaussian processes (KISS-GP) for low-dimensional inputs and KISS-GP with product kernels for high-dimensional inputs. The efficacy of DKL-KAN is evaluated in terms of computational training time and test prediction accuracy across a wide range of applications. Additionally, the effectiveness of DKL-KAN is also examined in modeling discontinuities and accurately estimating prediction uncertainty. The results indicate that DKL-KAN outperforms DKL-MLP on datasets with a low number of observations. Conversely, DKL-MLP exhibits better scalability and higher test prediction accuracy on datasets with large number of observations. 

\end{abstract}


\section{Introduction}
\label{sec:intro}
 
During the 1990s, before the emergence of deep learning, Gaussian Processes (GP) (\cite{williams2006gaussian})  became popular as alternatives to neural networks (NN) because of their ability to quantify uncertainty and their often superior performance. 
During this period, benchmarking datasets primarily consisted of tabular data, leading to the perception that representation learning could be entirely bypassed, focusing solely on the relationship between vectors in the input space. However, in the early 2010s, the community's focus shifted towards more intricate domains such as sequences (\cite{mikolov2010recurrent}), images (\cite{krizhevsky2012imagenet}), and speech (\cite{hinton2012deep}), driven by a series of works leveraging the inductive biases of domain-specific NNs. This shift posed a challenge for GPs, primarily due to the difficulty in tailoring kernels that express useful inductive biases for these diverse domains. In response to these challenges, researchers proposed methods for learning GP kernels directly from data (\cite{wilson2013gaussian}), aiming to improve the adaptability of GPs to diverse datasets by automatically capturing relevant features.

GP (\cite{williams2006gaussian}) are favored in Bayesian modeling due to their transparent  interpretability and reliable uncertainty quantification. Typically, these models involve a few kernel hyperparameters optimized with respect to the marginal likelihood.
Nonetheless, for many common choices, the kernel remains fixed, restricting the capability of GP to learn representations from the data that could improve predictions, thereby functioning primarily as smoothing mechanisms.. 
This restricts the applicability of GP to data that is high-dimensional and highly structured.
Conversely, deep neural networks (DNN) (\cite{lecun2015deep}) are renowned for learning strong representations which are utilized for predicting unseen test inputs. Despite their good performance, deterministic NN tend to produce overconfident predictions (\cite{guo2017calibration}) and lack trustworthy uncertainty estimates. The Bayesian approach to NN aims to mitigate these challenges; however, despite recent progress in variational inference and sampling techniques, inference in Bayesian NN remains difficult due to complicated posteriors and the huge number of parameters. Additionally, Bayesian NNs generally require multiple forward passes to obtain several samples of the predictive posterior to average over. Consequently, it is logical to combine the uncertainty representation advantages of GPs with the representation-learning advantages of NNs to leverage the strengths of both. This gave rise to deep kernel learning (DKL) (\cite{wilson2016deep}) which uses an NN to extract meaningful representations that are subsequently fed into a standard GP, allowing for end-to-end learning of all model parameters. Both kernel hyperparameters and network parameters can be jointly trained by using varational inference or maximizing the marginal likelihood. In \cite{wilson2016deep}, DKL has shown to outperform conventional shallow kernel learning methods, such as the radial basis function (RBF) kernel and standard NNs on a wide range of applications.

Multilayer perceptrons (MLP) (\cite{haykin1998neural}), a cornerstone of modern deep learning, consist of multiple layers of nodes, including an input layer, one or more hidden layers, and an output layer. These networks are capable of learning complex non-linear relationships within data, as evidenced by the universal approximation theorem (\cite{hornik1989multilayer}). Despite their success, MLP are often criticized for their lack of interpretability \cite{cranmer2023interpretable}, susceptibility to overfitting, scalablity, and issues related to vanishing or exploding gradients during training.

Kolmogorov-Arnold Networks (KAN) (\cite{liu2024kan}) have garnered significant attention lately as a novel neural network architecture aimed at addressing some of the inherent limitations of MLP. The foundation of KAN lies in the Kolmogorov-Arnold representation theorem, which asserts that any multivariate continuous function can be expressed as a finite sum of univariate functions and addition operations. This theorem, initially proven by Kolmogorov in 1957 to solve Hilbert's 13th problem (\cite{kolmogorov1957representation, braun2009constructive}), laid the groundwork for various neural network designs that leverage its decomposition principles.
Unlike MLP, which rely on linear weights and fixed activation functions, KAN use learnable activation functions, typically implemented as splines or B-splines. This design choice allows KANs to adaptively shape their activation functions to better fit the underlying data, thereby improving both accuracy and interpretability. 

The main contributions of this paper are as follows. First, we introduce a scalable deep kernel using KAN (DKL-KAN) as an alternative to the traditional DKL using MLP (DKL-MLP). 
Additionally, we present a comparative analysis of two DKL-KAN configurations: one matching the neuron and layer count of DKL-MLP, and another with a similar number of trainable parameters, ensuring a rigorous evaluation against the baseline. Our method is tested across various benchmark UCI regression datasets (\cite{uci_datasets}), focusing on both computational training time and prediction accuracy. Additionally, we also investigate the efficacy of DKL-KAN in modeling discontinuities and accurately calculating uncertainty bounds. To address the challenges of large datasets, we use Kernel Interpolation for Scalable Structured Gaussian Processes (KISS-GP) (\cite{wilson2015kernel}) for low input dimensions and KISS-GP with scalable kernel interpolation for product kernels (SKIP) (\cite{gardner2018product}) for high-dimensional inputs. 
The remainder of the paper is organized as follows. Section \ref{sec:related_works} discusses the related works. Section \ref{sec:kan} delves into the brief explanation of KAN. This is followed by explanation of GP in Section \ref{sec:gp}. The findings of this study are detailed in Section \ref{sec:experiments}, and finally, the conclusions are succinctly presented in Section \ref{sec:conclusions}

\section{Related Works}
\label{sec:related_works}

\subsection{Gaussian Process}
\label{subsec:gaussian_process}

GPs, traditional nonparametric modeling tools and NNs, known for their flexibility in fitting complex data patterns, have an intriguing equivalence (\cite{neal1996priors, lee2017deep, matthews2018gaussian, garriga2018deep, novak2018bayesian, yang2019scaling}). As derived by \cite{neal1996priors}, in the limit of infinite width, a one-layer NN corresponds to a GP. This equivalence implies that for infinitely wide networks, an i.i.d. prior over weights and biases can be replaced with a corresponding GP prior over functions, allowing exact Bayesian inference for regression using NNs. 

The concept of DKL was first presented by \cite{wilson2016deep}, building upon the framework of KISS-GP (\cite{wilson2015kernel}).
This approach combines deep NN with traditional covariance functions using local kernel interpolation to create complicated kernels. Although DKL excels in approximating intricate kernels for GP, it exhibits notable drawbacks. It does not support stochastic gradient training and is inadequate for classification tasks. In this regard, \cite{wilson2016stochastic} introduced the stochastic variational deep kernel learning  approach, which allows the model in performing classification tasks, multi-task learning, and stochastic gradient optimization. \cite{calandra2016manifold} extended the general framework by introducing manifold GP, which involve applying a kernel function to a parameterized transformation of the input data. This transformation can be parameterized using NN. The study of intricate kernel learning has also been a focus within certain specialized fields. For instance, \cite{al2017learning} used a recurrent neural network to acquire kernels characterized by recurrent patterns. In contrast, \cite{achituve2021personalized} explored using NN to learn a shared kernel function within the context of personalized federated learning. There has also been initiatives aimed at the regularization of models based on DKL. \cite{liu2021deep} introduced a latent-variable framework that integrates a stochastic encoding of inputs to allow regularized representation learning. Recently, \cite{achituve2023guided} introduced an innovative methodology for training DKL. This method uses the uncertainty quantification from a GP with an infinite-width deep NN kernel to steer the optimization of DKL.

The flexibility of deep NN can however lead to overfitting in DKL models, as identified by \cite{ober2021promises}. They discovered that DKL often correlates all input data points in an effort to minimize the effect of the complexity penalty term in the marginal likelihood. This challenge was tackled by using full Bayesian inference over the network parameters using Markov Chain Monte Carlo (MCMC) techniques. However, MCMC is inefficient when dealing with high-dimensional posteriors. To address this, \cite{matiasamortized} has introduced amortized variational DKL, which leverages deep NN to generate embeddings for the kernel function and to learn variational distributions over inducing points, thereby reducing overfitting by locally smoothing predictions.

\subsection{Kolmogorov-Arnold Networks}
\label{subsec:kan}

Recent research has proposed several extensions and variants of KANs to improve their performance and reduce computational complexity. For instance, Fourier KAN (\cite{xu2024fourierkan}), Wavelet KAN (\cite{bozorgasl2024wav}), RBF-KAN (\cite{li2024kolmogorov}),  fractional KAN (\cite{aghaei2024fkan}), BSRBF-KAN (\cite{ta2024bsrbf}), rKAN (\cite{aghaei2024rkan}) and sineKAN (\cite{reinhardt2024sinekan}) replace B-splines with other function bases to achieve more accurate solutions. Another notable variant is the Chebyshev KAN (\cite{ss2024chebyshev}), which use orthogonal polynomials as univariate functions. Each of these variants aims to address specific limitations or improve certain capabilities of the original KAN architecture. KANs have also shown promise in the context of solving partial differential equations and operator learning (\cite{liu2024kan, shukla2024comprehensive}). Reference \cite{wang2024kolmogorov} combined KANs with physics-informed neural networks (PINNs) to address the 2D Poisson equation, achieving superior performance compared to traditional methods. Similarly, \cite{abueidda2024deepokan} introduced DeepOKAN, an RBF-based KAN operator network, for solving 2D orthotropic elasticity problems. These studies highlighted the potential of KANs to tackle complex scientific computing tasks more effectively than MLPs. The efficacy of KANs has also been explored in various other applications, including computer vision (\cite{azam2024suitability}), medical image segmentation (\cite{li2024u}), and learning graphs (\cite{kiamari2024gkan}. However, despite these advancements, the scalability and computational efficiency of KANs remain significant challenges (\cite{liu2024kan}).

\section{Kolmogorov-Arnold Networks}
\label{sec:kan}

KANs represent an innovative class of neural networks that leverage the theoretical insights provided by the Kolmogorov-Arnold representation theorem. This theorem provides a theoretical basis for expressing any multivariate continuous function as a finite composition of univariate continuous functions and additions. By exploiting this theorem, KANs aim to overcome some of the limitations associated with traditional neural network architectures, particularly in terms of function approximation and interpretability. This theorem, also known as the superposition theorem, is a fundamental result in approximation theory. It asserts that any continuous multivariate function on a bounded domain can be represented as a composition of a finite number of univariate functions and linear operations. Assuming a smooth known function $f: [0,1]^{k} \rightarrow \mathbb{R}$, continous univariate functions $\phi_{i,j}: [0,1] \rightarrow \mathbb{R}$, and continous functions $\psi_j: \mathbb{R} \rightarrow \mathbb{R}$, the theorem states:
\begin{equation}
f(\boldsymbol{x}) = \sum_{j=1}^{2k+1} \psi_j\left(\sum_{i=1}^{k} \phi_{i,j}\left(x_i\right)\right),
\end{equation}
for some $k$. Irrespective of the chosen functions, an $L$ layer deep KAN can be expressed in matrix form as follows:
\begin{equation*}
    \hat{f}(x) = (\Psi_{L-1} \circ \cdots \circ \Psi_1 \circ \Psi_0)(x),
\end{equation*}
where each $\Psi_{l}$ represents the function matrix of the $l^{th}$ KAN layer. If $n_l$ denotes the number of neurons in layer $l$, then the functional matrix $\Psi_l$ that links neurons in layer $l$ to those in layer $l+1$ is defined as follows:
\begin{equation*}
   \Psi_l=\left(\begin{array}{cccc}\phi_{1,1}(\cdot) & \phi_{1,2}(\cdot) & \cdots & \phi_{1, n_l}(\cdot) \\ \phi_{2,1}(\cdot) & \phi_{2,2}(\cdot) & \cdots & \phi_{2, n_l}(\cdot) \\ \vdots & \vdots & \vdots & \vdots \\ \phi_{n_{l+1}, 1}(\cdot) & \phi_{n_{l+1}, 2}(\cdot) & \cdots & \phi_{n_{l+1}, n_l}(\cdot)\end{array}\right) ,
\end{equation*}
where $\phi_{i,j}$ signifies the activation function connecting neuron $i$ in layer $l$ to neuron $j$ in layer $l+1$. Let $\mathcal{D}$ denote the derivative operator and $\epsilon$ be a constant based on the behaviour of $f(.)$. The convergence of this approximation is characterized by:
\begin{equation*}
\max _{|\gamma| \leq m} \sup _{x \in[0,1]^k}\left\|\mathcal{D}^\gamma\left\{\hat{f}(x) - f(x)\right\}\right\|<\epsilon.
\end{equation*}
For more details, please refer to \cite{liu2024kan}.


The first KAN paper (\cite{liu2024kan}) introduced the residual activation function $\phi(x)$, which is defined as the sum of the base function $b(x) = \text{silu}(x)$ and the spline function $s(x)$, each multiplied by their respective weight matrices $w_b$ and $w_s$. This spline function $s(x)$ is represented as a linear combination of B-splines $B_i(x)$:
\begin{equation*}
    s(x) = \sum_i c_i B_i(x),
\end{equation*}
where $c_i$ are trainable constants. The activation function is configured with $w_s=1$ and $s(x) \approx 0$, and $w_b$ is initialized using Xavier initialization. The residual activation function is then given by:
\begin{equation*}
    \phi(x) = w_b b(x) + w_s s(x).
\end{equation*}

Due to the high computational cost of KANs, a more efficient variant was introduced in \cite{efficient-kan}. This version follows the methodology of \cite{liu2024kan} but optimizes the computations using B-spline basis functions combined linearly, thereby reducing memory usage and simplifying the computational process. The authors also replaced the incompatible L1 regularization on input samples with L1 regularization on weights, incorporated learnable scales for activation functions, and changed the initialization of the base weight and spline scaler matrices to Kaiming uniform initialization. These modifications significantly improved performance on the MNIST dataset. In this paper, we utilize this efficient KAN for comparison with MLP.
\section{Gaussian Processes}
\label{sec:gp}

GP is a probabilistic non-parametric approach which defines a prior over functions and can be utilized for regression tasks. A GP is exhaustively described by its mean function, $m(\bm{x})$, and a covariance function known as the kernel function, $k(\bm{x}, \bm{x}')$. Importantly, for any finite set of inputs, a GP will output a multivariate Gaussian distribution. In the context of GPR, the underlying assumption is that the function to be approximated is a sample from a GP. The mean function, $m(\bm{x})$, and covariance function, $k(\bm{x}, \bm{x}')$, are represented by the equations:
\begin{align*}
m(\bm{x}) &= \mathbb{E}[f(\bm{x})], \\
k(\bm{x}, \bm{x}') &= \mathbb{E}[(f(\bm{x}) - m(\bm{x}))(f(\bm{x}') - m(\bm{x}'))].
\end{align*}
In practice, it is common to use a constant mean function, i.e., $m(\bm{x}) = c$, and focus on defining the covariance function. The covariance function, measures the similarity between data points. One common choice is the standard radial basis function (RBF) kernel. Assuming $\sigma^2$ as the signal variance, $l_d$ representing the length scales for each dimension, and $D$ indicating the number of dimensions, we have:
\begin{equation*}
k(\bm{x}, \bm{x}') = \sigma^2 \exp\left(-\sum_{d=1}^D \frac{(x_d - x'_d)^2}{2l_d^2}\right).
\end{equation*}
The length scales $l_d$ control the relevance of different dimensions. When $l_d$ is small, changes in the $d$-th dimension have a large effect on the covariance, making that dimension more relevant. 

Given a dataset $\mathcal{D}$ of $n$ input vectors, $X=\left\{\mathbf{x}_1, \ldots, \mathbf{x}_n\right\}$, each having dimension $D$, these index an $n \times 1$ vector of targets $\mathbf{y}=\left(y\left(\mathbf{x}_1\right), \ldots, y\left(\mathbf{x}_n\right)\right)^{\top}$. Assuming the presence of additive Gaussian noise, the relationship between the target $y(\mathbf{x})$ and a function $f(\mathbf{x})$ can be modeled as:
\begin{equation*}
    y(\mathbf{x}) \mid f(\mathbf{x}) \sim \mathcal{N}\left(y(\mathbf{x}) ; f(\mathbf{x}), \sigma_{y}^2\right),
\end{equation*}
where $\sigma_y^2$ is the variance of the noise. Assuming all covariance matrices implicitly rely on the kernel hyperparameters $\theta$, we have the predictive distribution of the GP at the $n_{*}$ test points indexed by $X_{*}$ to be expressed as:
$$
\begin{gathered}
\mathbf{f}_* \mid X_*, X, \mathbf{y}, \boldsymbol{\theta}, \sigma_{y}^2 \sim \mathcal{N}\left(\mathbb{E}\left[\mathbf{f}_*\right], \operatorname{cov}\left(\mathbf{f}_*\right)\right), \\
\mathbb{E}\left[\mathbf{f}_*\right]=m(X_{*})+k_{\theta}(X_{*},X)\left[k_{\theta}(X,X)+\sigma_{y}^2 I\right]^{-1} \mathbf{y} \\
\operatorname{cov}\left(\mathbf{f}_*\right)=k_{\theta}(X_*,X_*)-k_{\theta}(X_*,X)\left[k_{\theta}(X,X)+\sigma_{y}^2 I\right]^{-1} k_{\theta}(X,X_*),
\end{gathered}
$$
where,
$$
\mathbf{f}_*=f_*(X)=\left[f_*\left(\mathbf{x}_1\right), \ldots, f_*\left(\mathbf{x}_n\right)\right]^{\top}.
$$
Here, $k_{\theta}(X_*,X)$ is an $n_* \times n$ matrix of covariances between the GP evaluated at $X_*$ and $X$. Additionally, $m(X_*)$ designates the $n_* \times 1$ mean vector, and $k_{\theta}(X,X)$ represents the $n \times n$ covariance matrix evaluated at training inputs $X$. 
The conditional probability density of the target variable $\mathbf{y}$, given the parameters $\theta$ and input data $X$, can be expressed as:
\begin{equation*}
\log p(\mathbf{y} \mid \theta, X) \propto -\left[\mathbf{y}^{\top}\left(k_{\theta}(X,X) + \sigma^2 I\right)^{-1} \mathbf{y} + \log \left|k_{\theta}(X,X) + \sigma^2 I\right|\right].
\end{equation*}

\subsection{Deep kernel}

A deep kernel integrates a neural network with a traditional kernel function, improving the model's ability to capture complex patterns in data. 
The neural network effectively compresses and distills the input space, mapping it to a lower-dimensional representation that is more informative for the GP model. 
The transformation induced by the neural network can allow the model to fit more complicated, non-smooth, and anisotropic functions, unlike the standard kernels like RBF, which model smooth and isotropic functions.
Also, by mapping a high-dimensional input space to a lower-dimensional one using a neural network, we can mitigate the curse of dimensionality. This makes the GPR more tractable and can improve its generalization by focusing on the most informative dimensions of the data.

\begin{figure}[htbp]
    \centering
    \includegraphics[width=0.95\linewidth]{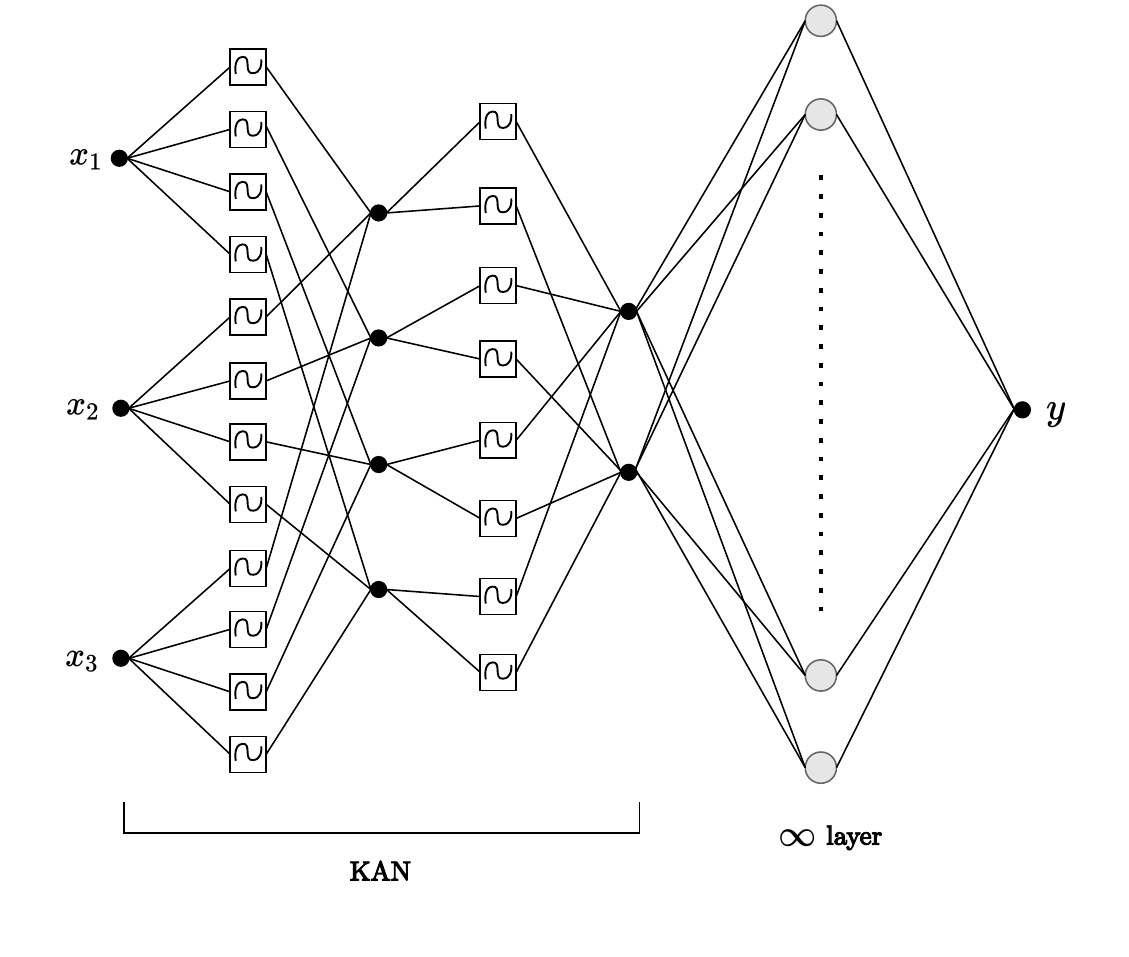}
    \caption{Schematic of DKL-KAN with 3 inputs and 1 output}
    \label{fig:gp_kan_schematic}
\end{figure}

Assuming $\phi(\bm{x};\bm{\beta})$ is a nonlinear mapping given by a neural network parameterized by $\bm{\beta}$, and $k(\bm{x}_i, \bm{x}_j' \mid \alpha)$ representing the standard RBF kernel parameterized by $\alpha$, we transform the inputs as:
\begin{equation*}
    k(\bm{x}_i,\bm{x}_j'\mid \alpha) \rightarrow k(\phi(\bm{x}_i; \bm{\beta}), \phi(\bm{x}_j'; \bm{\beta})|\alpha, \bm{\beta}).
\end{equation*}
We investigate the effectiveness of using KANs for DKL in this study. Fig. \ref{fig:gp_kan_schematic} shows the schematic of DKL-KAN with three inputs and one output.
We simultaneously optimize all deep kernel hyperparameters, denoted as $\gamma = (\alpha, \beta)$, by maximizing the log marginal likelihood $\mathcal{L}$ of the exact GP. To compute the derivatives of $\mathcal{L}$ with respect to $\gamma$, we use the following chain rule:
\begin{equation*}
    \frac{\partial \mathcal{L}}{\partial \boldsymbol{\alpha}}=\frac{\partial \mathcal{L}}{\partial  k_{\gamma}(X,X)} \frac{\partial k_{\gamma}(X,X)}{\partial \boldsymbol{\alpha}}, \quad \frac{\partial \mathcal{L}}{\partial \mathbf{\beta}}=\frac{\partial \mathcal{L}}{\partial  k_{\gamma}(X,X)} \frac{\partial  k_{\gamma}(X,X)}{\partial \phi(\bm{x}_i; \bm{\beta})} \frac{\partial\phi(\bm{x}_i; \bm{\beta}))}{\partial \beta} .
\end{equation*}

The implicit derivative of the log marginal likelihood with respect to covariance matrix  $k_{\gamma}(X,X)$ is expressed as
\begin{equation*}
    \frac{\partial \mathcal{L}}{\partial k_{\gamma}(X,X)}=\frac{1}{2}\left(k_{\gamma}(X,X)^{-1} \mathbf{y} \mathbf{y}^{\top} k_{\gamma}(X,X)^{-1} - k_{\gamma}(X,X)^{-1}\right),
\end{equation*}
where the noise covariance $\sigma^2 I$ is included within the covariance matrix and is considered part of the base kernel hyperparameters $\alpha$.

A significant computational challenge in DKL is resolving the linear system $\left(k_{\gamma}(X, X) + \sigma^2 I\right)^{-1} \mathbf{y}$. Additionally, within the realm of kernel learning, evaluating the logarithm of the determinant $\log \left|k_{\gamma}(X, X) + \sigma^2 I\right|$ as part of the marginal likelihood presents another difficulty. The conventional method uses the Cholesky decomposition of the $n \times n$ matrix $k_{\gamma}(X, X)$, which requires $\mathcal{O}(n^3)$ operations and $\mathcal{O}(n^2)$ storage. After completing inference, the computation of the predictive mean entails a cost of $\mathcal{O}(n)$, while calculating the predictive variance incurs a cost of $\mathcal{O}(n^2)$ for each test point $\mathbf{x}_*$.

\subsection{Scaling to large datasets for low input dimensions}
\label{subsec:kiss_gp}

Due to the cubic scaling of computations with the number of data points in the previously mentioned approach, we use structural kernel interpolation (SKI) using local cubic kernel interpolation to construct GP models for large datasets (\cite{wilson2015kernel, wilson2015thoughts}). Many strategies to scale GP models for large datasets belong to the class of inducing point methods (\cite{snelson2005sparse, hensman2013gaussian, silverman1985some}). These methods replace the exact kernel with an approximation to accelerate computations. For $m$ inducing points, the approximate kernels lead to computational complexity of $\mathcal{O}\left(m^2 n + m^3\right)$ and storage complexity of $\mathcal{O}\left(m n + m^2\right)$ for GP inference and learning (\cite{quinonero2005unifying}), resulting in a predictive mean and variance cost of $\mathcal{O}(m)$ and $\mathcal{O}\left(m^2\right)$ per test case. 
Nevertheless, when using a limited number of inducing points, these techniques experience a considerable decline in accuracy (\cite{wilson2014covariance}). In contrast, methods that exploit structural properties, like Kronecker (\cite{saatcci2012scalable}) and Toeplitz (\cite{cunningham2008fast}), leverage the covariance matrix to achieve scalable inference and learning without relying on approximations, thereby preserving both scalability and predictive precision.
However, the necessity of an input grid limits the applicability of these methods to most problems. SKI (\cite{wilson2015kernel}) integrates the benefits of both inducing point and structure-exploiting methods, explaining how the accuracy and efficiency of an inducing point method depend on the number of inducing points, kernel choice, and interpolation method. By selecting different interpolation strategies for SKI, novel inducing point methods can be devised. KISS-GP (\cite{wilson2015kernel}) uses SKI with local cubic and inverse distance weight interpolation to create a sparse approximation to the cross-covariance matrix between inducing points and original training points. Hence to ensure scalability, we replace all occurrences of \( k_{\gamma}(X, X) \) with the KISS-GP covariance matrix
\begin{equation*}
    k_{\gamma}(X, X) \approx W k_{\gamma}(U, U) W^{\top},
\end{equation*}
where $W$ represents a sparse matrix of interpolation weights, containing only four non-zero elements per row for local cubic interpolation. The term $k_{\gamma}(U, U)$ is a covariance matrix derived from the deep kernel, evaluated over $m$ latent inducing points $U=\left[\mathbf{u}_i\right]_{i=1 \ldots m}$. The computational complexity for KISS-GP training is $\mathcal{O}(n + h(m))$, with  $h(m)$ being nearly linear in $m$. The ability to have $m \approx n$ enables KISS-GP to achieve near-exact accuracy in its approximation, maintaining a non-parametric representation while ensuring linear scaling in $n $ and $\mathcal{O}(1)$ time per test point prediction. This approach is especially efficient when the input dimensionality is less (generally less than four), as the cost of grid construction escalates exponentially with the amount of data.

\subsection{Scaling to large datasets for high input dimensions}
\label{subsec:skip}
To tackle the aforementioned problem, we use SKIP as introduced by \cite{gardner2018product}. SKIP is designed to handle large datasets with high-dimensional inputs efficiently. It comprises of two primary elements: SKI and the product kernel structure. The product kernel structure in SKIP mitigates the curse of dimensionality inherent in SKI. For data characterized by $d$ input dimensions and denoting $\otimes$ as element-wise multiplication, the kernel matrix within the product kernel structure is represented as:
\begin{equation*}
k_{\gamma}(X,X) = k_{\gamma}(X,X)^{(1)} \otimes \ldots \otimes k_{\gamma}(X,X)^{(d)}.
\end{equation*}
This formulation ensures that SKIP maintains linear time complexity, even when dealing with high dimensional inputs.

\section{Experiments}
\label{sec:experiments}

\subsection{Data normalization}
\label{subsec:data_normalization}

All the training datasets were normalized using empirical cumulative distribution function technique prior to the initiation of the training process. This normalization technique transforms data into a uniform distribution, which can be particularly useful if the original distribution is unknown or if the data does not follow a Gaussian distribution which is a common assumption in many machine learning algorithms. It is also more robust to outliers as it ranks the datapoints rather than directly scaling their values. 

\subsection{Results}
\label{subsec:results}

The GP models and deep kernel is programmed using GPyTorch library respectively with a PyTorch backend. We set the loss function as negative of exact marginal log likelihood and the optimizer used is Adam with tuned hyperparameters. We set the learning rate as 0.075 with an exponential decay rate of 0.997 for smooth convergence of the loss. The GP models are trained for a total of 2500 epochs with a patience of 1000 epochs. All training is performed on an NVIDIA A100-SXM4-40GB GPU (1 GPU with 512 GB RAM). The codes for this paper will be made available at \href{https://github.com/shrenikvz/dkl-kan}{https://github.com/shrenikvz/dkl-kan} upon publication.

\subsubsection{UCI regression datasets}
\label{subsubsection:uci_regression_datasets}

We evaluate our models on a diverse set of benchmark UCI regression problems with varying sizes and characteristics. The base kernel is fixed as the RBF kernel. For DKL approaches, we reduce the dimensionality to 2 before feeding it to the base kernel. The neural network architecture for DKL-MLP consists of three hidden layers with 1000, 500, and 50 neurons, respectively. To fairly compare DKL-KAN with DKL-MLP, we train two models using KANs: DKL-KAN1 and DKL-KAN2. DKL-KAN1 has the same number of neurons and layers as DKL-MLP, while DKL-KAN2 has approximately the same number of trainable parameters as DKL-MLP. DKL-KAN2's architecture includes three hidden layers with 256, 128, and 64 neurons, respectively. We utilize RMSE as the metric to evaluate the test prediction accuracy.

\begin{table}[htbp]
    \centering
    \caption{RMSE performance of different GP models on UCI regression datasets with n observations and d input dimensions. The results are averaged over 5 equal partitions (90\% train, 10\% test) of the data with one standard deviation.}
    \vspace{0.5cm}
    \resizebox{0.9\textwidth}{!}{%
    \begin{tabular}{cccccccc}
        \hline
        \rule{0pt}{2ex}
        Datasets & n & d & SKI/SKIP used (for scalability) & GP & DKL-MLP & DKL-KAN1 & DKL-KAN2 
        \rule{0pt}{2ex}\\
        \hline
        \rule{0pt}{2ex}
        Solar & 1066 & 10 & No & 1.07 ± 0.32 & 1.41 ± 0.36 & \textbf{1.04 ± 0.20} & 1.13 ± 0.31 \\
        Airfoil & 1503 & 5 & No & 2.83 ± 0.85 & 3.67 ± 0.67 &  \textbf{2.21 ± 0.42} & 2.5 ± 0.55 \\
        Wine & 1599 & 11 & No & 0.55 ± 0.08 & 0.63 ± 0.15 & \textbf{0.53 ± 0.07}  &  0.54 ± 0.09\\
        Gas & 2565 & 128 & No & 0.16 ± 0.04 & 0.15 ± 0.04 & \textbf{0.15 ± 0.03} & 0.15 ± 0.06 \\
        Skillcraft & 3338 & 19 & No & \textbf{0.27 ± 0.04} & 0.35 ± 0.04 & \textbf{0.27 ± 0.04} & 0.28 ± 0.04 \\
        SML & 4137 & 26 & No & 0.72 ± 0.08 & 1.18 ± 0.17 & \textbf{0.69 ± 0.11} & 0.88 ± 0.19 \\
        Parkinsons & 5875 & 20 & No & \textbf{3.85 ± 2.93} & 11.72 ± 10.78 & 6.27 ± 2.14 & 13.26 ± 11.54 \\
        Pumadyn & 8192 & 32 & No & \textbf{0.29 ± 0.00} & 1.2 ± 0.3 & 0.83 ± 0.26 & 1.52 ± 0.96 \\
        Elevators & 16599 & 18 & No & 0.14 ± 0.01 & 0.32 ± 0.27 & \textbf{0.13 ± 0.04} & 0.15 ± 0.02 \\
        Kin40k & 40000 & 8 & Yes & 0.32 ± 0.03 & \textbf{0.29 ± 0.05} & 0.30 ± 0.06 & 0.33 ± 0.08 \\
        Protein & 45730 & 9 & Yes & 0.47 ± 0.01 & \textbf{0.45 ± 0.02}& 0.46 ± 0.04 &  0.46 ± 0.03\\
        Tamielectric & 45781 & 3 & Yes & \textbf{0.29 ± 0.00} & 0.55 ± 0.09 & 0.59 ± 0.07 & 0.59 ± 0.14  \\
        KEGG & 48827 & 22 & Yes & 0.34 ± 0.23 & \textbf{0.30 ± 0.20} & 0.31 ± 0.21 & 0.33 ± 0.20\\
        Ctslice & 53500 & 385 & Yes & 5.14 ± 0.09 & 3.19 ± 0.12 & \textbf{3.11 ± 0.14} & 3.21 ± 0.19\\
        KEGGU & 63608 & 27 & Yes & 0.25 ± 0.17 & \textbf{0.24 ± 0.09} & 0.25 ± 0.11 & 0.26 ± 0.10\\
        \hline
    \end{tabular}%
    }
    \label{tab:rmse_gp}
\end{table}

\begin{table}[htbp]
    \centering
    \caption{Total number of trainable parameters of different GP models on UCI regression datasets}
    \label{tab:parameters_gp}
     \vspace{0.5cm}
    \resizebox{0.5\textwidth}{!}{%
    \begin{tabular}{ccccc}
        \hline
        \rule{0pt}{2ex}
        Datasets & GP & DKL-MLP & DKL-KAN1 & DKL-KAN2 
        \rule{0pt}{2ex}\\
        \hline
        \rule{0pt}{2ex}
        Solar & 13 & 536657 & 5351005 &  436485\\
        Airfoil & 8 & 531657 & 5301005 & 423685 \\
        Wine & 14 & 537657 & 5361005 & 439045 \\
        Gas & 131 & 654657 & 6531005 & 738565 \\
        Skillcraft  & 22 & 545657 & 5441005 & 459525 \\
        SML  & 29 & 552657 & 5511005 & 477445 \\
        Parkinsons & 23 & 546657 & 5451005 & 462085 \\
        Pumadyn & 35 & 558657 & 5571005 & 492805 \\
        Elevators & 21 & 544657 & 5431005 & 456965 \\
        Kin40k  & 11 & 534657 & 5331005 & 431365 \\
        Protein & 12 & 535657 & 5341005 & 433925 \\
        Tamielectric & 6 & 529657 & 5281005 & 418565 \\
        KEGG  & 23 & 546657 & 5451005 & 462085 \\
        Ctslice  & 388 & 911657 & 9101005 & 1396485 \\
        KEGGU  & 30 & 553657 & 5521005 & 480005 \\
        \hline
    \end{tabular}%
    }
\end{table}

\begin{table}[htbp]
    \centering
    \caption{Average computational training runtime (in s) of different GP models on UCI regression datasets (Compute: NVIDIA A100-SXM4-40GB chip (1 GPU with 512 GB RAM))}
    \label{tab:runtime_gp}
     \vspace{0.5cm}
    \resizebox{0.5\textwidth}{!}{%
    \begin{tabular}{ccccc}
        \hline
        \rule{0pt}{2ex}
        Datasets & GP & DKL-MLP & DKL-KAN1 & DKL-KAN2 
        \rule{0pt}{2ex}\\
        \hline
        \rule{0pt}{2ex}
        Solar & 9.83 & 17.13 & 33.6 & 25.19 \\
        Airfoil & 9 & 17.09 & 50.44 & 31.12 \\
        Wine & 10.63 & 17.97 & 47.6 & 32.7 \\
        Gas & 11.98 & 19.56 & 86.37 & 43.07 \\
        Skillcraft & 11.96 & 20.99 & 62.77 & 34.4 \\
        SML & 10.12 & 19.71 & 61.92 & 33.4 \\
        Parkinsons & 56.97 & 24.44 & 74.18 & 41.67 \\
        Pumadyn & 31.29 & 43.4 & 107.53 & 72.36 \\
        Elevators & 182.55 & 56.72 & 110.39 & 122.99 \\
        Kin40k & 190.39 &  100.98 & 122.47 & 140.22 \\
        Protein &  250.68 & 220.46 & 256.91 & 240.27 \\
        Tamielectric & 120.45  & 156.21  & 164.89 & 159.24 \\
        KEGG & 190.32 & 187.59 & 190.83 & 180.64 \\
        Ctslice & 206.38 & 201.54 & 257.27 & 203.71 \\
        KEGGU & 212.44 & 213.69 & 258.63 & 219.08 \\
        \hline
    \end{tabular}%
    }
\end{table}

\begin{table}[htbp]
    \centering
    \caption{RMSE performance of different GP models on UCI regression datasets with n observations and d input dimensions. The results are averaged over 20 equal partitions (90\% train, 10\% test) of the data with one standard deviation.}
    \vspace{0.5cm}
    \resizebox{0.9\textwidth}{!}{%
    \begin{tabular}{cccccccc}
        \hline
        \rule{0pt}{2ex}
        Datasets & n & d & SKI/SKIP used (for scalability) & GP & DKL-MLP & DKL-KAN1 & DKL-KAN2 
        \rule{0pt}{2ex}\\
        \hline
        \rule{0pt}{2ex}
        Kin40k & 40000 & 8 & No & 0.39 ± 0.04 &  0.42 ± 0.06 & \textbf{0.31 ± 0.10} &  0.35 ±  0.09\\
        Protein & 45730 & 9 & No &  0.53 ± 0.03 & 0.55 ± 0.05 & \textbf{0.51 ± 0.08} &  0.56 ± 0.09\\
        Tamielectric & 45781 & 3 & No & \textbf{0.29 ± 0.01} &  0.64 ± 0.11 &  0.62 ±  0.12 &  0.63 ± 0.12  \\
        KEGG & 48827 & 22 & No & 0.13 ± 0.02 & 0.11 ± 0.06 & \textbf{0.10 ± 0.06} & 0.12 ± 0.07\\
        Ctslice & 53500 & 385 & No & 6.21 ± 0.16 & 6.59 ± 0.23 &  \textbf{6.19 ± 0.05} &  6.43 ± 0.06 \\
        KEGGU & 63608 & 27 & No & 0.34 ± 0.15 & 0.34 ± 0.13 & \textbf{0.31 ± 0.17} & 0.33 ± 0.15 \\
        \hline
    \end{tabular}%
    }
    \label{tab:rmse_gp_1}
\end{table}

\begin{figure}
    \centering
    \includegraphics[width=0.8\linewidth]{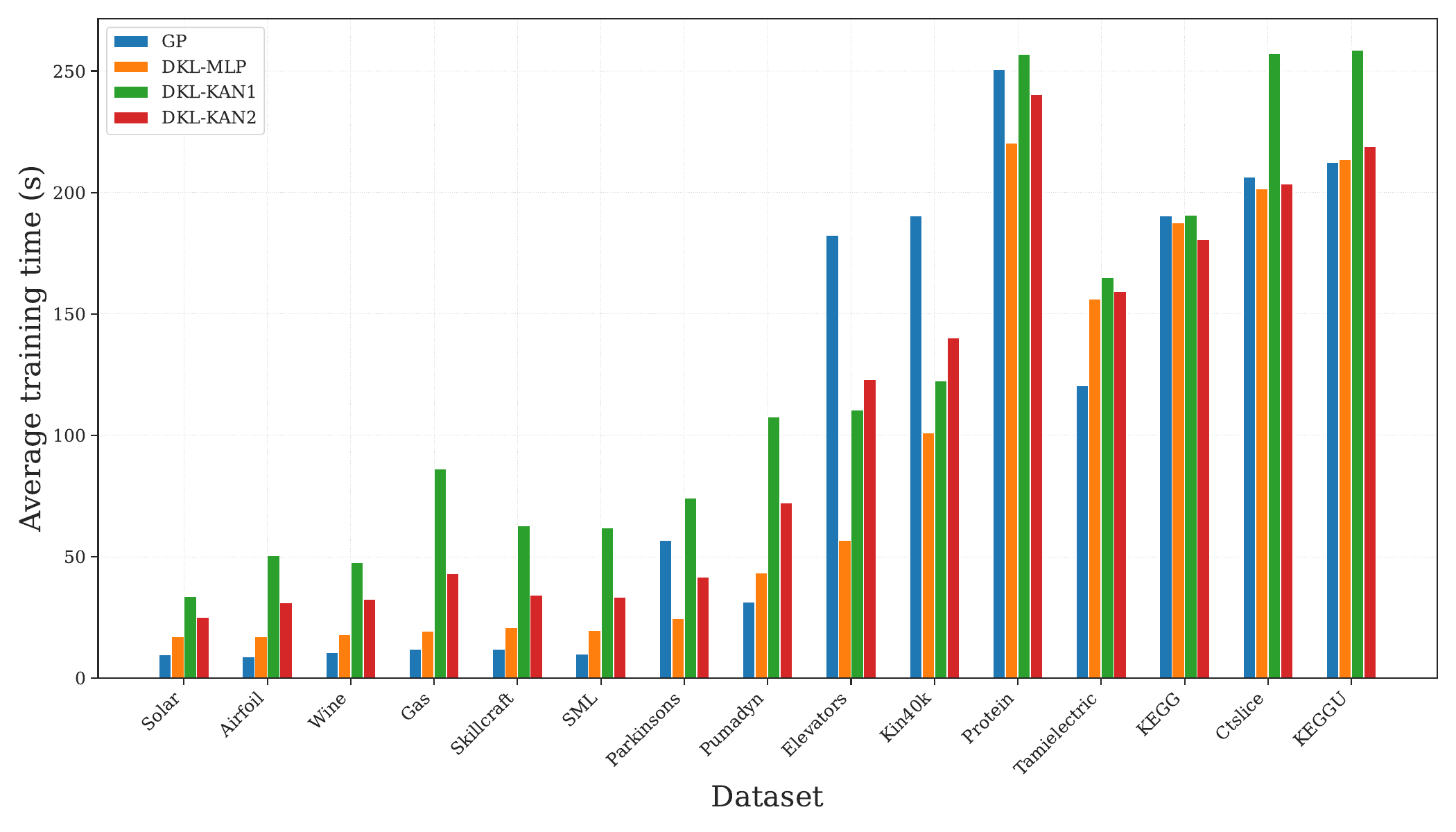}
    \caption{Average training time vs datasets (Compute: NVIDIA A100-SXM4-40GB chip (1 GPU with 512 GB RAM))}
    \label{fig:training_time_plot}
\end{figure}

Given that GP models could be trained in a finite time for datasets with fewer than 20,000 observations, we use SKI exclusively for datasets exceeding 20,000 observations. For these larger datasets, we use KISS-GP for DKL-MLP, DKL-KAN1, and DKL-KAN2, as the neural network maps to two output features, allowing for the use of SKI. Conversely, we use KISS-GP with product structure kernel (SKIP) for the standard GP, since the input features are not reduced before being fed to the base kernel.

As illustrated in Table \ref{tab:rmse_gp}, DKL-KAN1 shows superior performance on datasets with a low number of observations. Additionally, we observe that increasing the complexity of the KAN architecture leads to notable improvements in test prediction accuracy. However, for datasets with a large number of observations, DKL-MLP outperforms DKL-KAN, suggesting that the current iteration of KANs is less effective for large datasets.  However, we do see DKL-KAN1 outperform all other GP models for $\texttt{ctslice}$ dataset.

Table \ref{tab:rmse_gp_1} illustrates the RMSE performance of these GP models with an increased number of partitions (i.e., 20) for datasets with a large number of observations. Due to increased partitions (which means using lesser data to train on), we did not consider using SKI/SKIP for generating these results. It is evident that for these large datasets, increasing the number of partitions leads to better test prediction accuracy of DKL-KAN1 over DKL-MLP, further validating our earlier assertion that KANs in their current version are not scalable to large datasets.

There are also several instances where a standard GP shows relatively better performance than DKL. Table \ref{tab:parameters_gp} presents the total number of trainable parameters for these models, while Table \ref{tab:runtime_gp} provides the average computational training time (in seconds) for each model on specific datasets. Given that we used efficient KANs in this study, the training time of DKL-KANs is comparable to that of other GP models (Fig. \ref{fig:training_time_plot}).

\subsubsection{Step function}
\label{subsubsection:step_function}

\begin{figure}[htbp]
  \centering
  \begin{subfigure}[t]{0.33\textwidth}
  \centering
  \includegraphics[width=\textwidth]{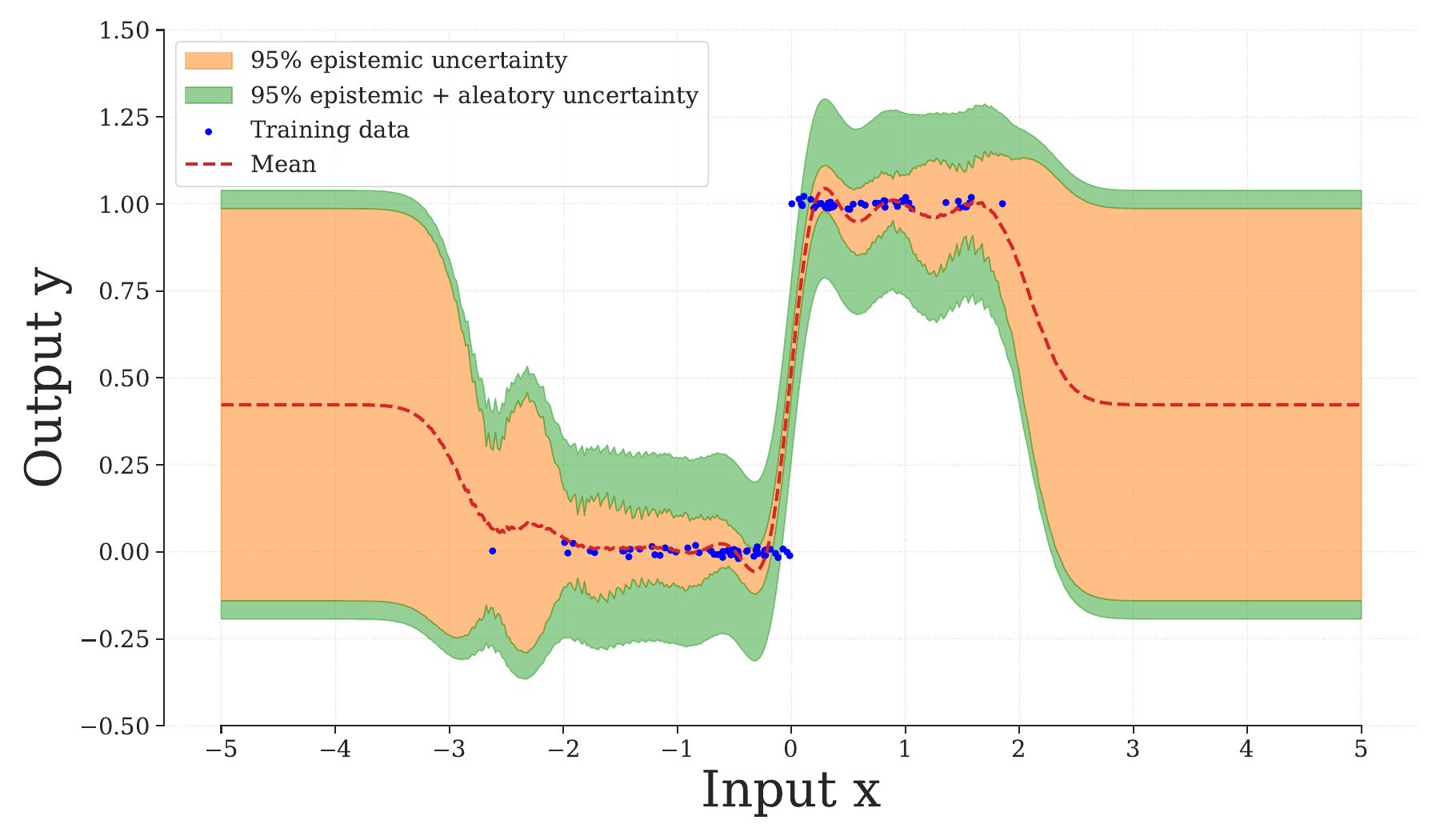}
   \subcaption{GP}
  \label{fig:step_rbf}
  \end{subfigure} \hfill
  \begin{subfigure}[t]{0.33\textwidth}
  \centering
  \includegraphics[width=\textwidth]{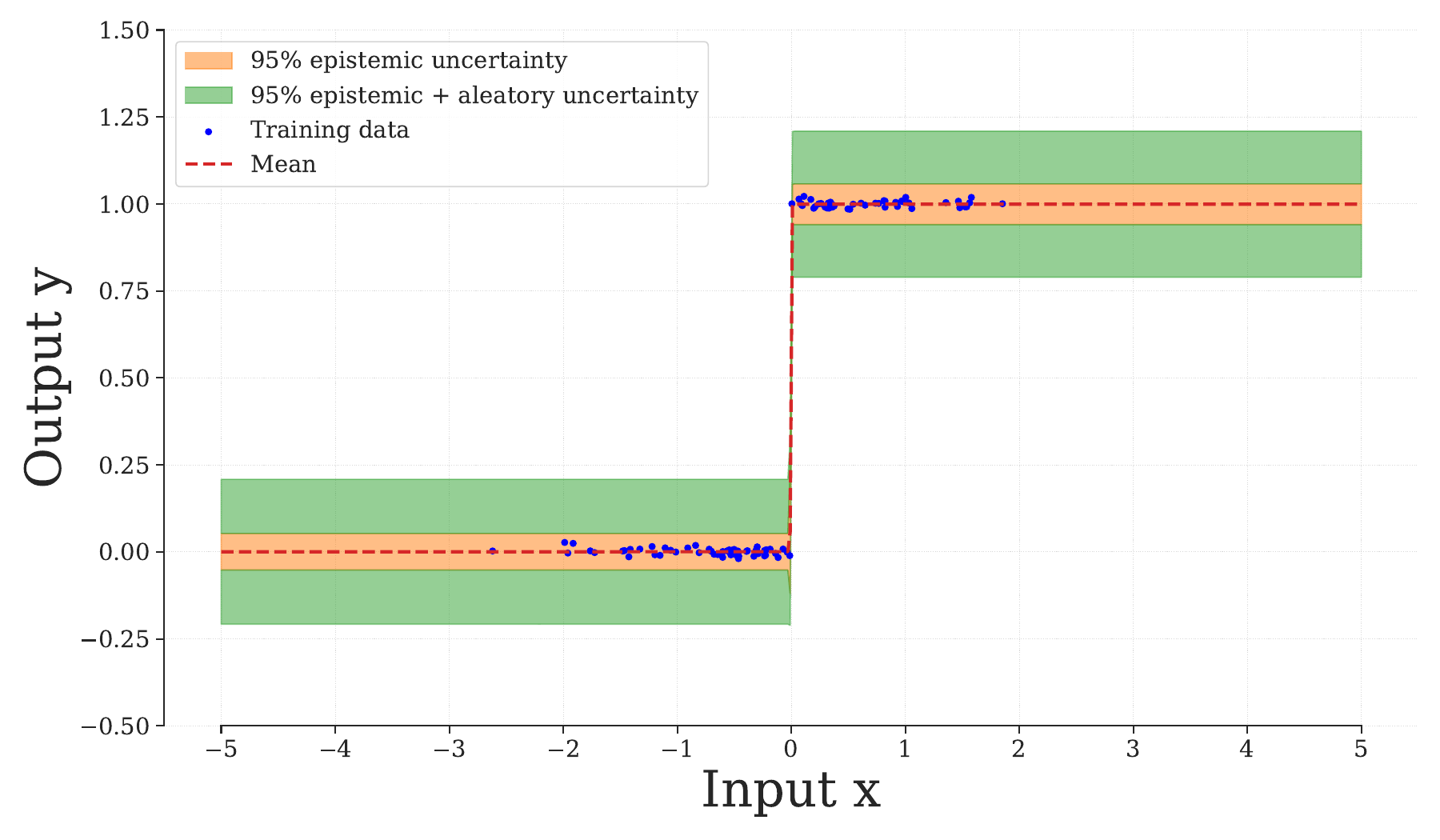}
   \subcaption{DKL-MLP}
  \label{fig:step_mlp}
  \end{subfigure} \hfill
  \begin{subfigure}[t]{0.33\textwidth}
  \centering
  \includegraphics[width=\textwidth]{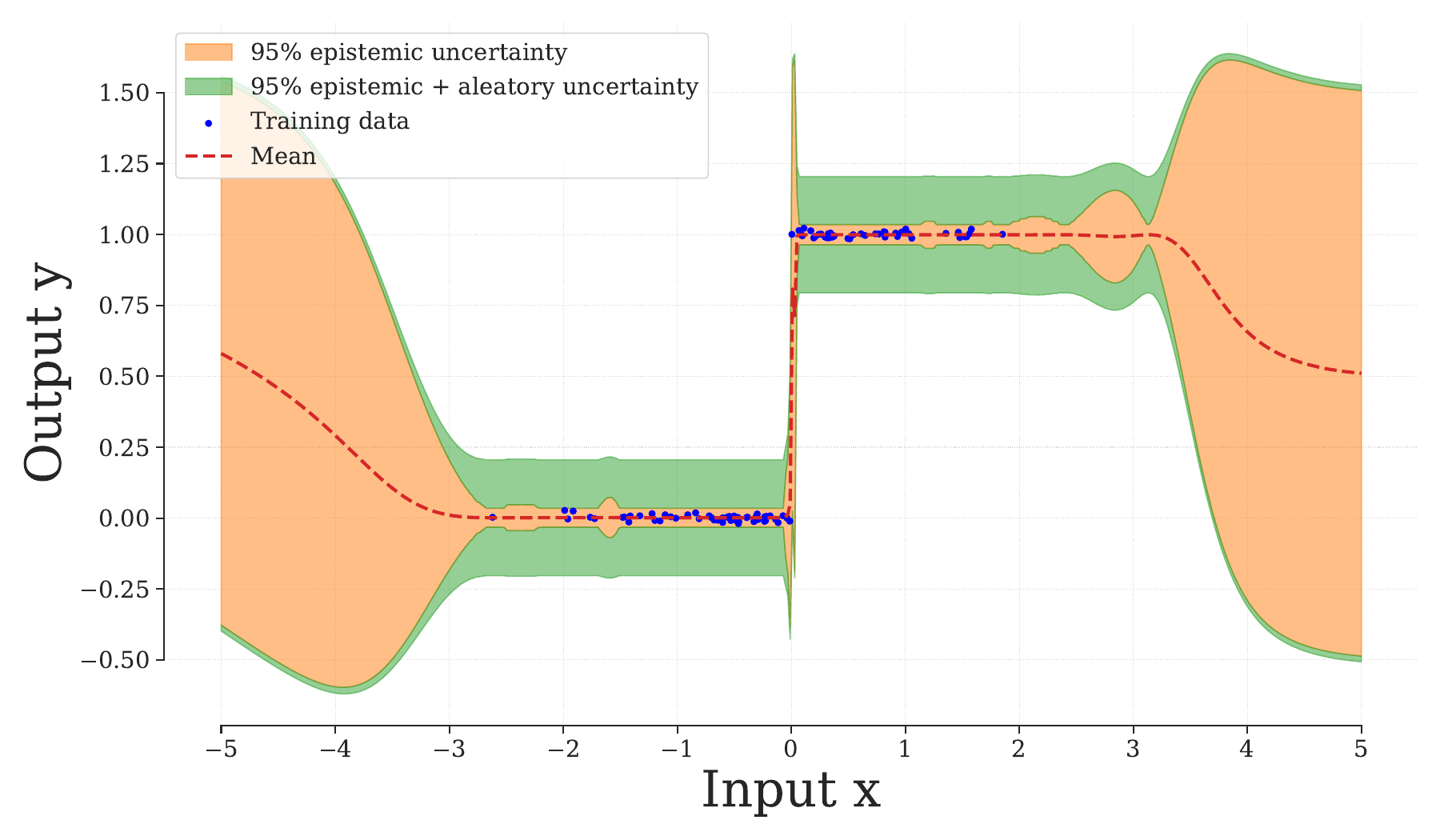}
   \subcaption{DKL-KAN}
  \label{fig:step_kan}
  \end{subfigure}
  \caption{GP prediction}
  \label{fig:gp_prediction}
\end{figure}

\begin{figure}[htbp]
  \centering
  \begin{subfigure}[t]{0.45\textwidth}
  \centering
  \includegraphics[width=\textwidth]{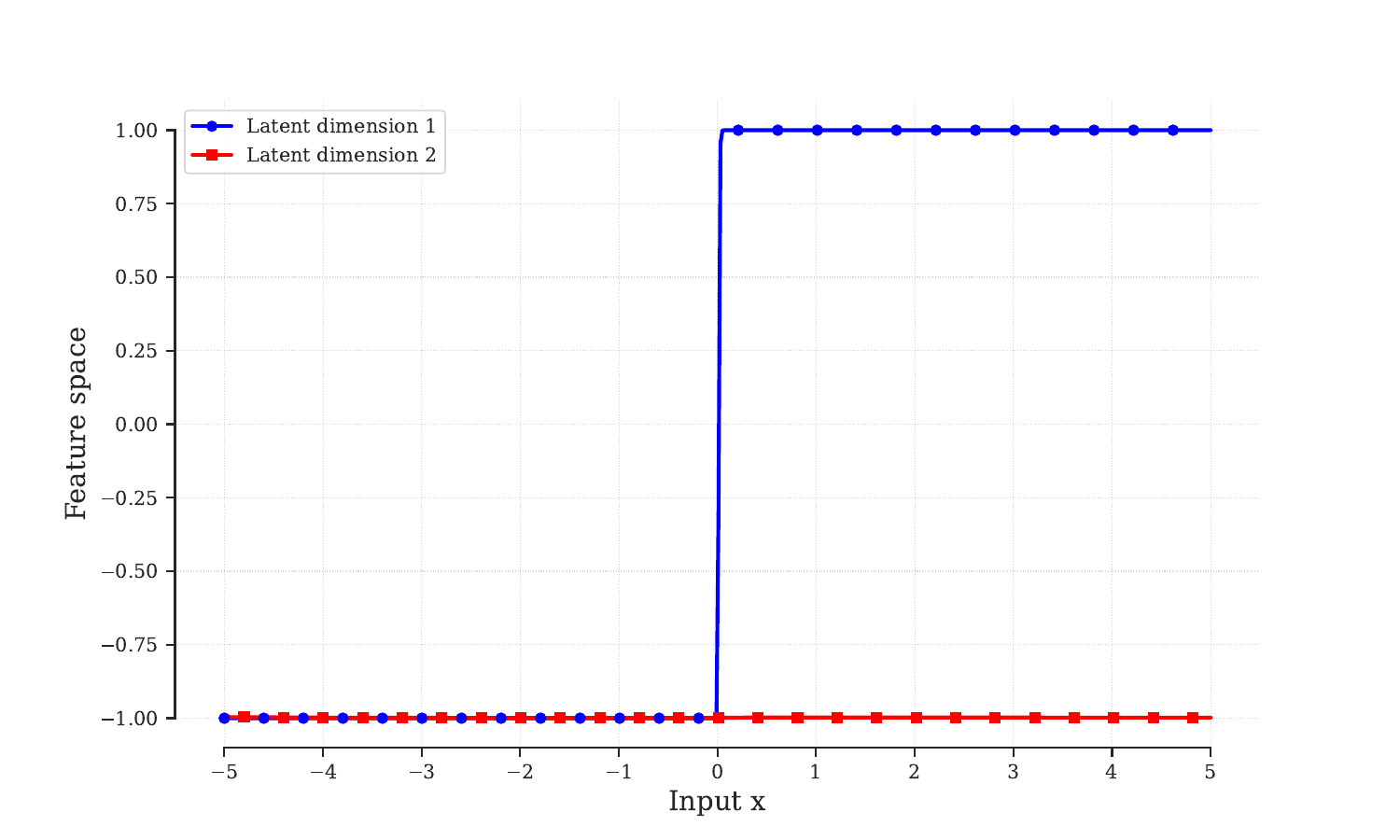}
   \subcaption{DKL-MLP}
  \label{fig:latent_mlp}
  \end{subfigure} \hfill
  \begin{subfigure}[t]{0.45\textwidth}
  \centering
  \includegraphics[width=\textwidth]{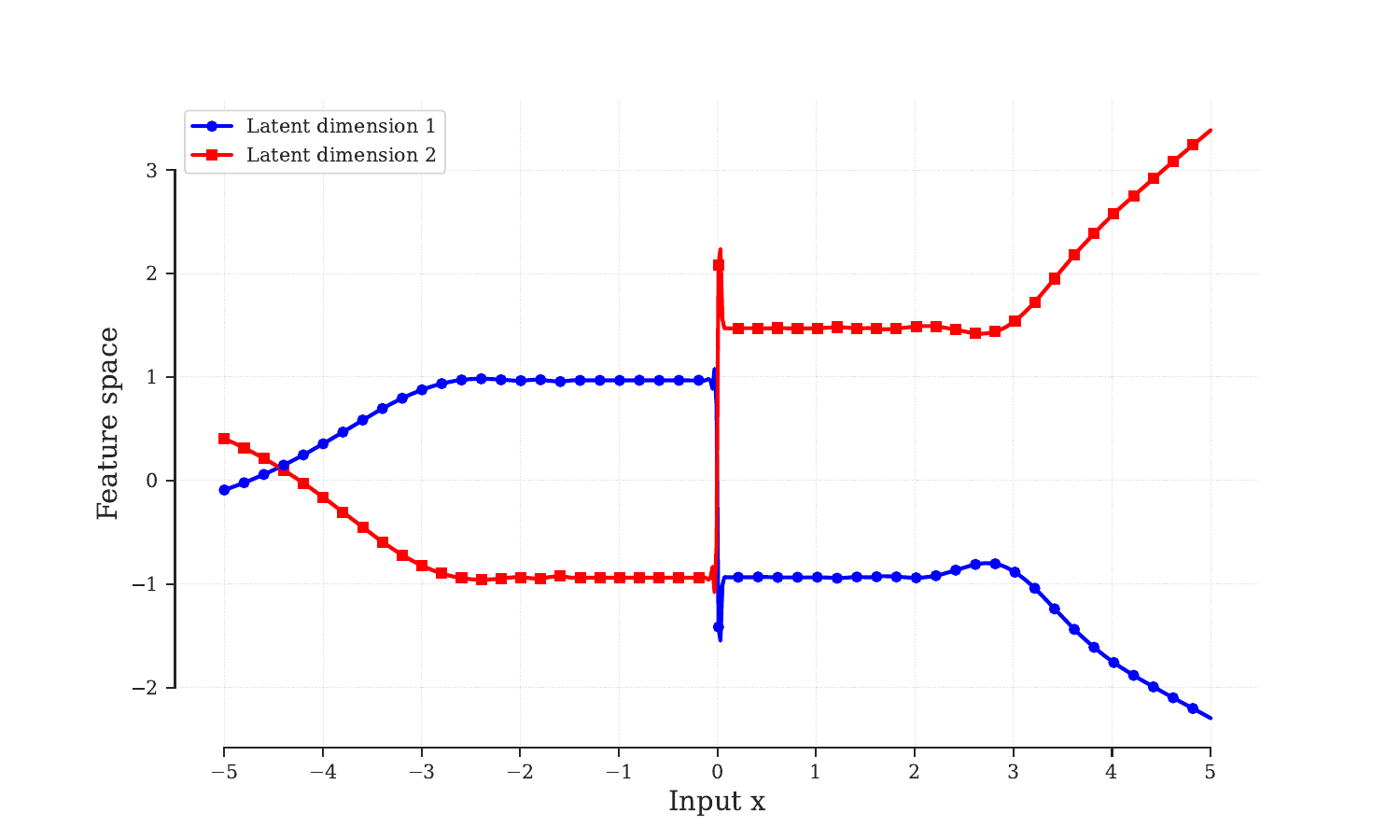}
   \subcaption{DKL-KAN}
  \label{fig:latent_kan}
  \end{subfigure}
  \caption{Learned mapping}
  \label{fig:latent_dimension}
\end{figure}

To evaluate the effectiveness of DKL-KAN in handling discontinuities and accurately estimating prediction uncertainty, we consider the step function depicted below
\begin{equation}
\begin{aligned}
y & = f(x) + w, \quad w \sim \mathcal{N}\left(0,0.01^2\right), \\
f(x) & = \begin{cases}0 & \text { if } x \leq 0 \\ 1 & \text { if } x > 0\end{cases}
\end{aligned}
\end{equation}
For the training process, 100 input points are sampled from a standard normal distribution, whereas the test set consists of 500 datapoints uniformly distributed between $-5$ and $5$. Given that we are modeling a step function, a neural network with a single hidden layer of 6 neurons is used for both DKL-MLP and DKL-KAN, with the reduced latent dimension set to 2. Fig. \ref{fig:gp_prediction} illustrates the predictive mean and uncertainty bands when different GP variants are tested on the test set. We observe a relative decrease in epistemic uncertainty in regions where training datapoints are present for GP, DKL-MLP, and DKL-KAN. The standard GP, which assumes a smooth function structure, struggles to model the discontinuity effectively. Due to DKL's joint supervised learning of input transformation and regression, the neural network captures the discontinuity in the reduced 2D feature space for both MLP and KAN (Fig.\ref{fig:latent_dimension}), facilitating a smooth mapping from reduced feature space to output that the GP can handle (\cite{calandra2016manifold}). Additionally, we observe that DKL-MLP maintains low uncertainty across the entire function domain as explained in \cite{calandra2016manifold}. In contrast, DKL-KAN exhibits increasing epistemic uncertainty in regions where training data is not present. The behavior of DKL-MLP with respect to uncertainty bounds is not ideal, as, in an optimal scenario, epistemic uncertainty should rise in areas lacking training data. DKL-KAN, however, offers a balanced solution by accurately capturing discontinuities and appropriately estimating prediction uncertainties.

\section{Conclusions}
\label{sec:conclusions}

In this study, we investigate the effectiveness of incorporating KANs into DKL. To ensure a fair comparison with DKL-MLP, we train two GP models using KANs: one with an identical number of neurons and layers as the MLP (DKL-KAN1) and another with approximately same number of trainable parameters (DKL-KAN2). Our findings indicate a relatively better performance of KANs when using a larger number of neurons for a fixed number of hidden layers. Moreover, DKL-KANs demonstrate superior performance compared to other GP models on datasets with a low number of observations. Conversely, DKL-MLP exhibit better scalability and higher test prediction accuracy on datasets with a large number of observations. This suggests that, in their current form, KANs are less effective for training on larger datasets. Additionally, DKL-KAN accurately captures discontinuities and prediction uncertainties, particularly in regions lacking training data. The ability of DKL-KAN to increase epistemic uncertainty where data is sparse ensures a more reliable model performance in practical applications.

\section*{Acknowledgements}

This work is supported by RTX Technology Research Center.

\bibliographystyle{unsrtnat}
\bibliography{references}

\begin{thebibliography}{54}
\providecommand{\natexlab}[1]{#1}
\providecommand{\url}[1]{\texttt{#1}}
\expandafter\ifx\csname urlstyle\endcsname\relax
  \providecommand{\doi}[1]{doi: #1}\else
  \providecommand{\doi}{doi: \begingroup \urlstyle{rm}\Url}\fi

\bibitem[Williams and Rasmussen(2006)]{williams2006gaussian}
Christopher~KI Williams and Carl~Edward Rasmussen.
\newblock \emph{Gaussian processes for machine learning}, volume~2.
\newblock MIT press Cambridge, MA, 2006.

\bibitem[Mikolov et~al.(2010)Mikolov, Karafi{\'a}t, Burget, Cernock{\`y}, and
  Khudanpur]{mikolov2010recurrent}
Tomas Mikolov, Martin Karafi{\'a}t, Lukas Burget, Jan Cernock{\`y}, and Sanjeev
  Khudanpur.
\newblock Recurrent neural network based language model.
\newblock In \emph{Interspeech}, volume~2, pages 1045--1048. Makuhari, 2010.

\bibitem[Krizhevsky et~al.(2012)Krizhevsky, Sutskever, and
  Hinton]{krizhevsky2012imagenet}
Alex Krizhevsky, Ilya Sutskever, and Geoffrey~E Hinton.
\newblock Imagenet classification with deep convolutional neural networks.
\newblock \emph{Advances in neural information processing systems}, 25, 2012.

\bibitem[Hinton et~al.(2012)Hinton, Deng, Yu, Dahl, Mohamed, Jaitly, Senior,
  Vanhoucke, Nguyen, Sainath, et~al.]{hinton2012deep}
Geoffrey Hinton, Li~Deng, Dong Yu, George~E Dahl, Abdel-rahman Mohamed, Navdeep
  Jaitly, Andrew Senior, Vincent Vanhoucke, Patrick Nguyen, Tara~N Sainath,
  et~al.
\newblock Deep neural networks for acoustic modeling in speech recognition: The
  shared views of four research groups.
\newblock \emph{IEEE Signal processing magazine}, 29\penalty0 (6):\penalty0
  82--97, 2012.

\bibitem[Wilson and Adams(2013)]{wilson2013gaussian}
Andrew Wilson and Ryan Adams.
\newblock Gaussian process kernels for pattern discovery and extrapolation.
\newblock In \emph{International conference on machine learning}, pages
  1067--1075. PMLR, 2013.

\bibitem[LeCun et~al.(2015)LeCun, Bengio, and Hinton]{lecun2015deep}
Yann LeCun, Yoshua Bengio, and Geoffrey Hinton.
\newblock Deep learning.
\newblock \emph{nature}, 521\penalty0 (7553):\penalty0 436--444, 2015.

\bibitem[Guo et~al.(2017)Guo, Pleiss, Sun, and Weinberger]{guo2017calibration}
Chuan Guo, Geoff Pleiss, Yu~Sun, and Kilian~Q Weinberger.
\newblock On calibration of modern neural networks.
\newblock In \emph{International conference on machine learning}, pages
  1321--1330. PMLR, 2017.

\bibitem[Wilson et~al.(2016{\natexlab{a}})Wilson, Hu, Salakhutdinov, and
  Xing]{wilson2016deep}
Andrew~Gordon Wilson, Zhiting Hu, Ruslan Salakhutdinov, and Eric~P Xing.
\newblock Deep kernel learning.
\newblock In \emph{Artificial intelligence and statistics}, pages 370--378.
  PMLR, 2016{\natexlab{a}}.

\bibitem[Haykin(1998)]{haykin1998neural}
Simon Haykin.
\newblock \emph{Neural networks: a comprehensive foundation}.
\newblock Prentice Hall PTR, 1998.

\bibitem[Hornik et~al.(1989)Hornik, Stinchcombe, and
  White]{hornik1989multilayer}
Kurt Hornik, Maxwell Stinchcombe, and Halbert White.
\newblock Multilayer feedforward networks are universal approximators.
\newblock \emph{Neural networks}, 2\penalty0 (5):\penalty0 359--366, 1989.

\bibitem[Cranmer(2023)]{cranmer2023interpretable}
Miles Cranmer.
\newblock Interpretable machine learning for science with pysr and
  symbolicregression. jl.
\newblock \emph{arXiv preprint arXiv:2305.01582}, 2023.

\bibitem[Liu et~al.(2024)Liu, Wang, Vaidya, Ruehle, Halverson,
  Solja{\v{c}}i{\'c}, Hou, and Tegmark]{liu2024kan}
Ziming Liu, Yixuan Wang, Sachin Vaidya, Fabian Ruehle, James Halverson, Marin
  Solja{\v{c}}i{\'c}, Thomas~Y Hou, and Max Tegmark.
\newblock Kan: Kolmogorov-arnold networks.
\newblock \emph{arXiv preprint arXiv:2404.19756}, 2024.

\bibitem[Kolmogorov(1957)]{kolmogorov1957representation}
Andrei~Nikolaevich Kolmogorov.
\newblock On the representation of continuous functions of many variables by
  superposition of continuous functions of one variable and addition.
\newblock In \emph{Doklady Akademii Nauk}, volume 114, pages 953--956. Russian
  Academy of Sciences, 1957.

\bibitem[Braun and Griebel(2009)]{braun2009constructive}
J{\"u}rgen Braun and Michael Griebel.
\newblock On a constructive proof of kolmogorov’s superposition theorem.
\newblock \emph{Constructive approximation}, 30:\penalty0 653--675, 2009.

\bibitem[Evans(2021)]{uci_datasets}
Trefor Evans.
\newblock Uci datasets.
\newblock \url{https://github.com/treforevans/uci_datasets}, 2021.
\newblock Accessed: 2024-06-20.

\bibitem[Wilson and Nickisch(2015)]{wilson2015kernel}
Andrew Wilson and Hannes Nickisch.
\newblock Kernel interpolation for scalable structured gaussian processes
  (kiss-gp).
\newblock In \emph{International conference on machine learning}, pages
  1775--1784. PMLR, 2015.

\bibitem[Gardner et~al.(2018)Gardner, Pleiss, Wu, Weinberger, and
  Wilson]{gardner2018product}
Jacob Gardner, Geoff Pleiss, Ruihan Wu, Kilian Weinberger, and Andrew Wilson.
\newblock Product kernel interpolation for scalable gaussian processes.
\newblock In \emph{International Conference on Artificial Intelligence and
  Statistics}, pages 1407--1416. PMLR, 2018.

\bibitem[Neal and Neal(1996)]{neal1996priors}
Radford~M Neal and Radford~M Neal.
\newblock Priors for infinite networks.
\newblock \emph{Bayesian learning for neural networks}, pages 29--53, 1996.

\bibitem[Lee et~al.(2017)Lee, Bahri, Novak, Schoenholz, Pennington, and
  Sohl-Dickstein]{lee2017deep}
Jaehoon Lee, Yasaman Bahri, Roman Novak, Samuel~S Schoenholz, Jeffrey
  Pennington, and Jascha Sohl-Dickstein.
\newblock Deep neural networks as gaussian processes.
\newblock \emph{arXiv preprint arXiv:1711.00165}, 2017.

\bibitem[Matthews et~al.(2018)Matthews, Rowland, Hron, Turner, and
  Ghahramani]{matthews2018gaussian}
Alexander G de~G Matthews, Mark Rowland, Jiri Hron, Richard~E Turner, and
  Zoubin Ghahramani.
\newblock Gaussian process behaviour in wide deep neural networks.
\newblock \emph{arXiv preprint arXiv:1804.11271}, 2018.

\bibitem[Garriga-Alonso et~al.(2018)Garriga-Alonso, Rasmussen, and
  Aitchison]{garriga2018deep}
Adri{\`a} Garriga-Alonso, Carl~Edward Rasmussen, and Laurence Aitchison.
\newblock Deep convolutional networks as shallow gaussian processes.
\newblock \emph{arXiv preprint arXiv:1808.05587}, 2018.

\bibitem[Novak et~al.(2018)Novak, Xiao, Lee, Bahri, Yang, Hron, Abolafia,
  Pennington, and Sohl-Dickstein]{novak2018bayesian}
Roman Novak, Lechao Xiao, Jaehoon Lee, Yasaman Bahri, Greg Yang, Jiri Hron,
  Daniel~A Abolafia, Jeffrey Pennington, and Jascha Sohl-Dickstein.
\newblock Bayesian deep convolutional networks with many channels are gaussian
  processes.
\newblock \emph{arXiv preprint arXiv:1810.05148}, 2018.

\bibitem[Yang(2019)]{yang2019scaling}
Greg Yang.
\newblock Scaling limits of wide neural networks with weight sharing: Gaussian
  process behavior, gradient independence, and neural tangent kernel
  derivation.
\newblock \emph{arXiv preprint arXiv:1902.04760}, 2019.

\bibitem[Wilson et~al.(2016{\natexlab{b}})Wilson, Hu, Salakhutdinov, and
  Xing]{wilson2016stochastic}
Andrew~G Wilson, Zhiting Hu, Russ~R Salakhutdinov, and Eric~P Xing.
\newblock Stochastic variational deep kernel learning.
\newblock \emph{Advances in neural information processing systems}, 29,
  2016{\natexlab{b}}.

\bibitem[Calandra et~al.(2016)Calandra, Peters, Rasmussen, and
  Deisenroth]{calandra2016manifold}
Roberto Calandra, Jan Peters, Carl~Edward Rasmussen, and Marc~Peter Deisenroth.
\newblock Manifold gaussian processes for regression.
\newblock In \emph{2016 International joint conference on neural networks
  (IJCNN)}, pages 3338--3345. IEEE, 2016.

\bibitem[Al-Shedivat et~al.(2017)Al-Shedivat, Wilson, Saatchi, Hu, and
  Xing]{al2017learning}
Maruan Al-Shedivat, Andrew~Gordon Wilson, Yunus Saatchi, Zhiting Hu, and Eric~P
  Xing.
\newblock Learning scalable deep kernels with recurrent structure.
\newblock \emph{Journal of Machine Learning Research}, 18\penalty0
  (82):\penalty0 1--37, 2017.

\bibitem[Achituve et~al.(2021)Achituve, Shamsian, Navon, Chechik, and
  Fetaya]{achituve2021personalized}
Idan Achituve, Aviv Shamsian, Aviv Navon, Gal Chechik, and Ethan Fetaya.
\newblock Personalized federated learning with gaussian processes.
\newblock \emph{Advances in Neural Information Processing Systems},
  34:\penalty0 8392--8406, 2021.

\bibitem[Liu et~al.(2021)Liu, Ong, Jiang, and Wang]{liu2021deep}
Haitao Liu, Yew-Soon Ong, Xiaomo Jiang, and Xiaofang Wang.
\newblock Deep latent-variable kernel learning.
\newblock \emph{IEEE Transactions on Cybernetics}, 52\penalty0 (10):\penalty0
  10276--10289, 2021.

\bibitem[Achituve et~al.(2023)Achituve, Chechik, and
  Fetaya]{achituve2023guided}
Idan Achituve, Gal Chechik, and Ethan Fetaya.
\newblock Guided deep kernel learning.
\newblock In \emph{Uncertainty in Artificial Intelligence}, pages 11--21. PMLR,
  2023.

\bibitem[Ober et~al.(2021)Ober, Rasmussen, and van~der Wilk]{ober2021promises}
Sebastian~W Ober, Carl~E Rasmussen, and Mark van~der Wilk.
\newblock The promises and pitfalls of deep kernel learning.
\newblock In \emph{Uncertainty in Artificial Intelligence}, pages 1206--1216.
  PMLR, 2021.

\bibitem[Matias et~al.()Matias, Mattos, Gomes, and Mesquita]{matiasamortized}
Alan~LS Matias, C{\'e}sar~Lincoln Mattos, Jo{\~a}o Paulo~Pordeus Gomes, and
  Diego Mesquita.
\newblock Amortized variational deep kernel learning.
\newblock In \emph{Forty-first International Conference on Machine Learning}.

\bibitem[Xu et~al.(2024)Xu, Chen, Li, Yang, Wang, Hu, and
  Ngai]{xu2024fourierkan}
Jinfeng Xu, Zheyu Chen, Jinze Li, Shuo Yang, Wei Wang, Xiping Hu, and Edith C-H
  Ngai.
\newblock Fourierkan-gcf: Fourier kolmogorov-arnold network--an effective and
  efficient feature transformation for graph collaborative filtering.
\newblock \emph{arXiv preprint arXiv:2406.01034}, 2024.

\bibitem[Bozorgasl and Chen(2024)]{bozorgasl2024wav}
Zavareh Bozorgasl and Hao Chen.
\newblock Wav-kan: Wavelet kolmogorov-arnold networks.
\newblock \emph{arXiv preprint arXiv:2405.12832}, 2024.

\bibitem[Li(2024)]{li2024kolmogorov}
Ziyao Li.
\newblock Kolmogorov-arnold networks are radial basis function networks.
\newblock \emph{arXiv preprint arXiv:2405.06721}, 2024.

\bibitem[Aghaei(2024{\natexlab{a}})]{aghaei2024fkan}
Alireza~Afzal Aghaei.
\newblock fkan: Fractional kolmogorov-arnold networks with trainable jacobi
  basis functions.
\newblock \emph{arXiv preprint arXiv:2406.07456}, 2024{\natexlab{a}}.

\bibitem[Ta(2024)]{ta2024bsrbf}
Hoang-Thang Ta.
\newblock Bsrbf-kan: A combination of b-splines and radial basic functions in
  kolmogorov-arnold networks.
\newblock \emph{arXiv preprint arXiv:2406.11173}, 2024.

\bibitem[Aghaei(2024{\natexlab{b}})]{aghaei2024rkan}
Alireza~Afzal Aghaei.
\newblock rkan: Rational kolmogorov-arnold networks.
\newblock \emph{arXiv preprint arXiv:2406.14495}, 2024{\natexlab{b}}.

\bibitem[Reinhardt and Gleyzer(2024)]{reinhardt2024sinekan}
Eric~AF Reinhardt and Sergei Gleyzer.
\newblock Sinekan: Kolmogorov-arnold networks using sinusoidal activation
  functions.
\newblock \emph{arXiv preprint arXiv:2407.04149}, 2024.

\bibitem[SS(2024)]{ss2024chebyshev}
Sidharth SS.
\newblock Chebyshev polynomial-based kolmogorov-arnold networks: An efficient
  architecture for nonlinear function approximation.
\newblock \emph{arXiv preprint arXiv:2405.07200}, 2024.

\bibitem[Shukla et~al.(2024)Shukla, Toscano, Wang, Zou, and
  Karniadakis]{shukla2024comprehensive}
Khemraj Shukla, Juan~Diego Toscano, Zhicheng Wang, Zongren Zou, and George~Em
  Karniadakis.
\newblock A comprehensive and fair comparison between mlp and kan
  representations for differential equations and operator networks.
\newblock \emph{arXiv preprint arXiv:2406.02917}, 2024.

\bibitem[Wang et~al.(2024)Wang, Sun, Bai, Anitescu, Eshaghi, Zhuang, Rabczuk,
  and Liu]{wang2024kolmogorov}
Yizheng Wang, Jia Sun, Jinshuai Bai, Cosmin Anitescu, Mohammad~Sadegh Eshaghi,
  Xiaoying Zhuang, Timon Rabczuk, and Yinghua Liu.
\newblock Kolmogorov arnold informed neural network: A physics-informed deep
  learning framework for solving pdes based on kolmogorov arnold networks.
\newblock \emph{arXiv preprint arXiv:2406.11045}, 2024.

\bibitem[Abueidda et~al.(2024)Abueidda, Pantidis, and
  Mobasher]{abueidda2024deepokan}
Diab~W Abueidda, Panos Pantidis, and Mostafa~E Mobasher.
\newblock Deepokan: Deep operator network based on kolmogorov arnold networks
  for mechanics problems.
\newblock \emph{arXiv preprint arXiv:2405.19143}, 2024.

\bibitem[Azam and Akhtar(2024)]{azam2024suitability}
Basim Azam and Naveed Akhtar.
\newblock Suitability of kans for computer vision: A preliminary investigation.
\newblock \emph{arXiv preprint arXiv:2406.09087}, 2024.

\bibitem[Li et~al.(2024)Li, Liu, Li, Wang, Liu, and Yuan]{li2024u}
Chenxin Li, Xinyu Liu, Wuyang Li, Cheng Wang, Hengyu Liu, and Yixuan Yuan.
\newblock U-kan makes strong backbone for medical image segmentation and
  generation.
\newblock \emph{arXiv preprint arXiv:2406.02918}, 2024.

\bibitem[Kiamari et~al.(2024)Kiamari, Kiamari, and
  Krishnamachari]{kiamari2024gkan}
Mehrdad Kiamari, Mohammad Kiamari, and Bhaskar Krishnamachari.
\newblock Gkan: Graph kolmogorov-arnold networks.
\newblock \emph{arXiv preprint arXiv:2406.06470}, 2024.

\bibitem[Blealtan(2024)]{efficient-kan}
Blealtan.
\newblock Efficient kan.
\newblock \url{https://github.com/Blealtan/efficient-kan}, 2024.
\newblock Accessed: 2024-06-20.

\bibitem[Wilson et~al.(2015)Wilson, Dann, and Nickisch]{wilson2015thoughts}
Andrew~Gordon Wilson, Christoph Dann, and Hannes Nickisch.
\newblock Thoughts on massively scalable gaussian processes.
\newblock \emph{arXiv preprint arXiv:1511.01870}, 2015.

\bibitem[Snelson and Ghahramani(2005)]{snelson2005sparse}
Edward Snelson and Zoubin Ghahramani.
\newblock Sparse gaussian processes using pseudo-inputs.
\newblock \emph{Advances in neural information processing systems}, 18, 2005.

\bibitem[Hensman et~al.(2013)Hensman, Fusi, and Lawrence]{hensman2013gaussian}
James Hensman, Nicolo Fusi, and Neil~D Lawrence.
\newblock Gaussian processes for big data.
\newblock \emph{arXiv preprint arXiv:1309.6835}, 2013.

\bibitem[Silverman(1985)]{silverman1985some}
Bernhard~W Silverman.
\newblock Some aspects of the spline smoothing approach to non-parametric
  regression curve fitting.
\newblock \emph{Journal of the Royal Statistical Society: Series B
  (Methodological)}, 47\penalty0 (1):\penalty0 1--21, 1985.

\bibitem[Quinonero-Candela and Rasmussen(2005)]{quinonero2005unifying}
Joaquin Quinonero-Candela and Carl~Edward Rasmussen.
\newblock A unifying view of sparse approximate gaussian process regression.
\newblock \emph{The Journal of Machine Learning Research}, 6:\penalty0
  1939--1959, 2005.

\bibitem[Wilson(2014)]{wilson2014covariance}
Andrew~Gordon Wilson.
\newblock \emph{Covariance kernels for fast automatic pattern discovery and
  extrapolation with Gaussian processes}.
\newblock PhD thesis, University of Cambridge Cambridge, UK, 2014.

\bibitem[Saat{\c{c}}i(2012)]{saatcci2012scalable}
Yunus Saat{\c{c}}i.
\newblock \emph{Scalable inference for structured Gaussian process models}.
\newblock PhD thesis, Citeseer, 2012.

\bibitem[Cunningham et~al.(2008)Cunningham, Shenoy, and
  Sahani]{cunningham2008fast}
John~P Cunningham, Krishna~V Shenoy, and Maneesh Sahani.
\newblock Fast gaussian process methods for point process intensity estimation.
\newblock In \emph{Proceedings of the 25th international conference on Machine
  learning}, pages 192--199, 2008.

\end{thebibliography}

\end{document}